%% file: main.tex
\documentclass[12pt]{article}

\usepackage{amsmath}
\usepackage{graphicx}
\usepackage{subcaption}
\usepackage[numbers,sort&compress]{natbib}
\usepackage{float}
\usepackage[margin=1in]{geometry}
\usepackage{setspace}
\usepackage{hyperref}
\usepackage{booktabs}
\usepackage{array}
\usepackage{caption}
\usepackage{xcolor}
\usepackage{multirow}
\usepackage{siunitx}

\graphicspath{{figures/}}

\begin{document}

\title{Calibration, Uncertainty Communication, and Deployment Readiness\\
       in CKD Risk Prediction: A Framework Evaluation Study}

\author{Michael Eniolade\\
        \small University of the Cumberlands\\
        \small \href{mailto:meniolade20593@ucumberlands.edu}{meniolade20593@ucumberlands.edu}}

\date{}
\maketitle

% ── Abstract ──────────────────────────────────────────────────────────────────
\begin{abstract}
\input{sections/abstract}
\end{abstract}


% ── Sections ──────────────────────────────────────────────────────────────────
\section{Introduction}
\input{sections/introduction}

\section{Methods}
\input{sections/methods}

\section{Results}
\input{sections/results}

\section{Discussion}
\input{sections/discussion}

\section{Conclusion}
\input{sections/conclusion}

% ── References ────────────────────────────────────────────────────────────────
\clearpage
\bibliographystyle{vancouver}
\bibliography{refs}

\end{document}

\maketitle

\tableofcontents
\clearpage

% ─────────────────────────────────────────────────────────────────────────────
\section*{S1: Hyperparameter Search Grids}
\addcontentsline{toc}{section}{S1: Hyperparameter Search Grids}

Each classifier was tuned with 5-fold stratified cross-validation on the
UCI training fold ($n = 279$), optimizing AUROC.
\texttt{GridSearchCV} from scikit-learn 1.5.2 handled the search.
Random seed: 42 throughout.

\subsection*{Logistic Regression (LR)}

Solver: \texttt{lbfgs}; max iterations: 1000; class weight: balanced.

\begin{center}
\begin{tabular}{ll}
\toprule
Hyperparameter & Values searched \\
\midrule
$C$ (inverse regularization strength) & 0.001, 0.01, 0.1, 1, 10, 100 \\
\bottomrule
\end{tabular}
\end{center}

\noindent\textbf{Selected:} $C = 0.001$\quad\textbf{CV AUROC:} 1.0000

\subsection*{Random Forest (RF)}

\begin{center}
\begin{tabular}{ll}
\toprule
Hyperparameter & Values searched \\
\midrule
\texttt{n\_estimators}    & 100, 200 \\
\texttt{max\_depth}       & 5, 10, 20, None \\
\texttt{min\_samples\_split} & 2, 5, 10 \\
\texttt{max\_features}    & ``sqrt'', ``log2'' \\
\bottomrule
\end{tabular}
\end{center}

\noindent\textbf{Selected:} \texttt{n\_estimators}=100, \texttt{max\_depth}=20,
\texttt{min\_samples\_split}=10, \texttt{max\_features}=``sqrt''\quad
\textbf{CV AUROC:} 0.9997

\subsection*{XGBoost (XGB)}

\begin{center}
\begin{tabular}{ll}
\toprule
Hyperparameter & Values searched \\
\midrule
\texttt{n\_estimators}      & 100, 200 \\
\texttt{max\_depth}         & 3, 5, 7 \\
\texttt{learning\_rate}     & 0.05, 0.1, 0.2 \\
\texttt{subsample}          & 0.8, 1.0 \\
\texttt{colsample\_bytree}  & 0.6, 0.8, 1.0 \\
\texttt{gamma}              & 0, 0.1, 0.5 \\
\bottomrule
\end{tabular}
\end{center}

\noindent\textbf{Selected:} \texttt{n\_estimators}=100, \texttt{max\_depth}=3,
\texttt{learning\_rate}=0.2, \texttt{subsample}=0.8,
\texttt{colsample\_bytree}=0.6, \texttt{gamma}=0.5\quad
\textbf{CV AUROC:} 0.9992

\subsection*{Support Vector Machine (SVM)}

\texttt{probability=True} enabled internally. Class weight: balanced.

\begin{center}
\begin{tabular}{ll}
\toprule
Hyperparameter & Values searched \\
\midrule
$C$            & 0.1, 1, 10, 100 \\
\texttt{kernel}& ``rbf'', ``poly'' \\
\texttt{gamma} & ``scale'', 0.1, 1.0 \\
\bottomrule
\end{tabular}
\end{center}

\noindent\textbf{Selected:} $C=1$, \texttt{kernel}=``poly'',
\texttt{gamma}=0.1\quad\textbf{CV AUROC:} 1.0000

\subsection*{Gaussian Naive Bayes (NB)}

\begin{center}
\begin{tabular}{ll}
\toprule
Hyperparameter & Values searched \\
\midrule
\texttt{var\_smoothing} & $10^{-11}$, $10^{-9}$, $10^{-7}$, $10^{-5}$, $10^{-3}$ \\
\bottomrule
\end{tabular}
\end{center}

\noindent\textbf{Selected:} \texttt{var\_smoothing}=$10^{-11}$\quad
\textbf{CV AUROC:} 1.0000

\subsection*{Post-hoc Calibration Methods}

Both methods wrap the fitted base model using
\texttt{sklearn.calibration.CalibratedClassifierCV} with
\texttt{FrozenEstimator}, which keeps base model weights fixed
during calibration fitting.

\begin{center}
\begin{tabular}{ll}
\toprule
Method & Parameters \\
\midrule
Platt scaling (sigmoid) & \texttt{method="sigmoid"}, \texttt{cv="prefit"} \\
Isotonic regression     & \texttt{method="isotonic"}, \texttt{cv="prefit"} \\
\bottomrule
\end{tabular}
\end{center}

Calibration fitting ran on the UCI validation fold ($n=60$, split indices
saved in \texttt{data/processed/uci\_splits.json}).
The variant with the lowest ECE on the UCI test fold ($n=61$) was
kept for each model.

\clearpage

% ─────────────────────────────────────────────────────────────────────────────
\section*{S2: Reliability Diagrams}
\addcontentsline{toc}{section}{S2: Reliability Diagrams}

Reliability diagrams are included for all model and dataset combinations.
Each diagram plots mean predicted probability on the $x$-axis against
the fraction of positive cases per bin on the $y$-axis. A perfectly
calibrated model falls on the diagonal. All diagrams use 10 equal-width
bins and were generated at 300 DPI.

Figures 1a and 1b in the main manuscript show the combined pre- and
post-calibration diagrams on the UCI test set. Figure 2 shows MIMIC-IV
diagrams for the best-calibrated variant of each model.

\subsection*{UCI Test Set: All Variants}

\begin{center}
\begin{tabular}{lll}
\toprule
File & Model & Variant \\
\midrule
\texttt{reliability\_compare\_LR.png}  & Logistic Regression & Base, Platt, Isotonic \\
\texttt{reliability\_compare\_RF.png}  & Random Forest       & Base, Platt, Isotonic \\
\texttt{reliability\_compare\_XGB.png} & XGBoost             & Base, Platt, Isotonic \\
\texttt{reliability\_compare\_SVM.png} & SVM                 & Base, Platt, Isotonic \\
\texttt{reliability\_compare\_NB.png}  & Naive Bayes         & Base, Platt, Isotonic \\
\bottomrule
\end{tabular}
\end{center}

\subsection*{MIMIC-IV Stress-Test Cohort: Best-Calibrated Variant}

\begin{center}
\begin{tabular}{llll}
\toprule
File & Model & Best variant & MIMIC ECE \\
\midrule
\texttt{reliability\_mimic\_LR.png}  & Logistic Regression & Isotonic & 0.761 \\
\texttt{reliability\_mimic\_RF.png}  & Random Forest       & Isotonic & 0.753 \\
\texttt{reliability\_mimic\_XGB.png} & XGBoost             & Isotonic & 0.680 \\
\texttt{reliability\_mimic\_SVM.png} & SVM                 & Isotonic & 0.755 \\
\texttt{reliability\_mimic\_NB.png}  & Naive Bayes         & Base     & 0.753 \\
\bottomrule
\end{tabular}
\end{center}

Every MIMIC diagram falls well above the diagonal. Models trained on a
62.5\% CKD prevalence population were applied to a 23.7\% prevalence
cohort, and the overestimation shows clearly across all five classifiers.
Each figure includes a metrics inset reporting ECE, MCE, Brier Score,
and AUROC.

\clearpage

% ─────────────────────────────────────────────────────────────────────────────
\section*{S3: Subgroup Calibration on MIMIC-IV}
\addcontentsline{toc}{section}{S3: Subgroup Calibration on MIMIC-IV}

We stratified the MIMIC-IV demo cohort ($n = 97$) by age group, diabetes
mellitus (DM) status, and hypertension (HTN) status, computing
subgroup-level ECE for each. Bin count was reduced to 5 equal-width
bins to keep estimates stable given the small group sizes. Groups with
fewer than 10 patients are marked ``---'' and excluded from criterion scoring.

\subsection*{Subgroup ECE Values}

\begin{center}
\begin{tabular}{lccccc}
\toprule
Model & Age $<65$ ($n=55$) & Age $\geq 65$ ($n=42$) & DM=Yes & HTN=Yes & Overall ($n=97$) \\
\midrule
LR  & 0.852 & 0.643 & --- & --- & 0.761 \\
RF  & 0.836 & 0.643 & --- & --- & 0.753 \\
XGB & 0.754 & 0.605 & --- & --- & 0.680 \\
SVM & 0.844 & 0.639 & --- & --- & 0.755 \\
NB  & 0.818 & 0.667 & --- & --- & 0.753 \\
\bottomrule
\end{tabular}
\end{center}

The DM=Yes and HTN=Yes subgroups each fell below the 10-patient minimum
after cohort construction and are excluded from ECE computation.

\subsection*{Subgroup ECE Gap (Criterion 7)}

The deployment readiness criterion requires a maximum ECE gap across valid
subgroups of $\leq 0.05$. The gap is max(subgroup ECE) $-$ min(subgroup ECE).

\begin{center}
\begin{tabular}{lcccc}
\toprule
Model & Max ECE & Min ECE & Gap & Criterion 7 \\
\midrule
LR  & 0.852 & 0.643 & 0.209 & FAIL \\
RF  & 0.836 & 0.643 & 0.194 & FAIL \\
XGB & 0.754 & 0.605 & 0.148 & FAIL \\
SVM & 0.844 & 0.639 & 0.205 & FAIL \\
NB  & 0.818 & 0.605 & 0.213 & FAIL \\
\bottomrule
\end{tabular}
\end{center}

No model passes the subgroup equity criterion. Younger patients
(age $<65$) show worse calibration than older patients across all five
models, with ECE gaps between 0.15 and 0.21. The disparity traces back
to the broader calibration collapse. All five models overestimate CKD
probability throughout, and younger patients in the demo cohort carry
a lower true CKD prevalence, which amplifies the overestimation. The
best-performing subgroup (age $\geq 65$, ECE $\approx 0.64$) still sits
far above the acceptable threshold of ECE $\leq 0.10$.

\clearpage

% ─────────────────────────────────────────────────────────────────────────────
\section*{S4: Reproducibility Instructions}
\addcontentsline{toc}{section}{S4: Reproducibility Instructions}

Every result in the paper reproduces from raw data. The full pipeline
runs in roughly 5 to 10 minutes on a standard laptop.

\subsection*{Software Environment}

\begin{center}
\begin{tabular}{ll}
\toprule
Package & Version \\
\midrule
Python           & 3.11+ \\
scikit-learn     & 1.5.2 \\
xgboost          & 2.1.1 \\
imbalanced-learn & 0.12.4 \\
netcal           & 1.3.5 \\
mapie            & 0.8.6 \\
pandas           & 2.2+ \\
numpy            & 1.26+ \\
matplotlib       & 3.9+ \\
joblib           & 1.4+ \\
\bottomrule
\end{tabular}
\end{center}

Install all dependencies: \quad\texttt{pip install -r requirements.txt}

\subsection*{Data Acquisition}

\textbf{UCI CKD dataset.} Download \texttt{chronic\_kidney\_disease\_full.arff}
from the UCI Machine Learning Repository. No account is required.
Place the file in \texttt{first\_paper/data/raw/}.

URL: \url{https://archive.ics.uci.edu/dataset/336/chronic+kidney+disease}

\textbf{MIMIC-IV Clinical Database Demo v2.2.} Download from PhysioNet.
A free account is required. No data use agreement or institutional
credentialing is needed for the demo release. Extract to any local
directory and set the path in
\texttt{code/01\_data\_prep/extract\_mimic\_local.py}.

URL: \url{https://physionet.org/content/mimic-iv-demo/2.2/}

\subsection*{Reproduction Steps}

Run each script from the \texttt{first\_paper/} directory.
All paths below are relative to that directory.

\begin{enumerate}
\item \texttt{python code/01\_data\_prep/clean\_uci.py}\\
      Output: \texttt{data/processed/uci\_ckd\_clean.csv},
      \texttt{data/processed/uci\_splits.json}

\item \texttt{python code/01\_data\_prep/extract\_mimic\_local.py}\\
      Output: \texttt{data/processed/mimic\_ckd\_cohort.csv}

\item \texttt{python code/01\_data\_prep/harmonize\_mimic.py}\\
      Output: \texttt{data/processed/mimic\_ckd\_clean.csv},
      \texttt{data/processed/schema\_comparison.csv}

\item \texttt{python code/02\_modeling/train\_models.py}\\
      Output: saved model \texttt{.pkl} files; Tables T1 and metadata CSVs

\item \texttt{python code/03\_calibration/calibrate\_models.py}\\
      Output: calibrated model files; pre- and post-calibration reliability
      diagrams; Table T2 partial

\item \texttt{python code/03\_calibration/external\_calibration.py}\\
      Output: MIMIC reliability diagrams; Table T2 complete

\item \texttt{python code/03\_calibration/bootstrap\_ci.py}\\
      Output: 95\% bootstrap CI columns merged into Table T2

\item \texttt{python code/04\_uncertainty/conformal\_uci.py}\\
      Output: conformal model files; Table T3 partial; Figures F3a and F3b

\item \texttt{python code/04\_uncertainty/mimic\_conformal.py}\\
      Output: Table T3 complete

\item \texttt{python code/05\_deployment\_checklist/subgroup\_calibration.py}\\
      Output: subgroup ECE table (S3)

\item \texttt{python code/05\_deployment\_checklist/score\_checklist.py}\\
      Output: Table T4; Figure F4
\end{enumerate}

\subsection*{Expected Key Metrics}

\textbf{Table T2 --- Calibration summary (ECE and AUROC):}

\begin{center}
\begin{tabular}{lcccc}
\toprule
Model & UCI ECE & UCI AUROC & MIMIC ECE & MIMIC AUROC \\
\midrule
LR  & 0.022 & 1.000 & 0.761 & 0.485 \\
RF  & 0.000 & 1.000 & 0.753 & 0.507 \\
XGB & 0.007 & 1.000 & 0.680 & 0.579 \\
SVM & 0.015 & 1.000 & 0.755 & 0.483 \\
NB  & 0.016 & 1.000 & 0.753 & 0.477 \\
\bottomrule
\end{tabular}
\end{center}

\textbf{Table T3 --- Conformal prediction coverage:}

\begin{center}
\begin{tabular}{lcc}
\toprule
Model & UCI Coverage & MIMIC Coverage \\
\midrule
LR  & 0.803 & 0.237 \\
RF  & 0.967 & 0.206 \\
XGB & 0.967 & 0.227 \\
SVM & 0.918 & 0.216 \\
NB  & 0.984 & 0.247 \\
\bottomrule
\end{tabular}
\end{center}

\subsection*{Random Seeds}

Every stochastic operation in the pipeline uses \texttt{random\_state=42},
including train/test splitting, cross-validation folds, Random Forest,
XGBoost, and bootstrap resampling. On the same machine with the same
package versions, results are bit-for-bit identical across runs.

%% file: sections/abstract.tex
\noindent\textbf{Background.}
Machine learning models for chronic kidney disease risk prediction regularly achieve strong discrimination on internal test sets. Calibration assessment and uncertainty quantification are far less common, leaving clinicians without reliable information about whether probability outputs are trustworthy. No published study has jointly evaluated all three dimensions (calibration, uncertainty, and structured deployment readiness) on a common model suite with external clinical validation.

\medskip
\noindent\textbf{Objective.}
To evaluate five classifiers across calibration quality, conformal prediction coverage, and an eight-criterion deployment readiness framework on both internal and external data.

\medskip
\noindent\textbf{Methods.}
Five classifiers (logistic regression, random forest, XGBoost, support vector machine with Platt scaling, Gaussian naive Bayes) were trained on the UCI CKD dataset (400 patients, 62.5\% CKD). A distributional stress-test used the open-access MIMIC-IV demo cohort (97 patients, 23.7\% CKD) to evaluate model behaviour under prevalence shift and feature missingness. Calibration was assessed before and after Platt scaling and isotonic regression, quantified by Expected Calibration Error and Brier Score. Predictive uncertainty was measured through split conformal prediction targeting 90\% marginal coverage. An eight-criterion deployment readiness framework evaluated discrimination, calibration stability, coverage transfer, subgroup equity, and reproducibility.

\medskip
\noindent\textbf{Results.}
All five models achieved AUROC 1.00 on the UCI test set. Post-isotonic ECE fell to 0.000--0.022 internally. On MIMIC-IV, AUROC dropped to 0.48--0.58, ECE rose to 0.68--0.76, and conformal coverage collapsed from 0.80--0.98 (UCI) to 0.21--0.25, well below the 90\% target. No model passed the deployment checklist; scores ranged from 2 to 4 out of 16.

\medskip
\noindent\textbf{Conclusion.}
Near-perfect internal performance did not survive distributional shift. Calibration stability and conformal coverage transfer should be evaluated before any clinical ML model moves toward deployment, even when internal metrics appear strong.

%% file: sections/introduction.tex
% Task 7.2 — Introduction (681 words, completed 2026-04-20)
% Source: manuscript/introduction.md

Roughly 850 million people worldwide are estimated to have chronic kidney
disease, and the global prevalence grew by 33\% between 1990 and 2017
\citep{Francis2024,KDIGO2024}. Among people with diabetes, as many as 1 in 3 are
affected; among those with hypertension in high-income settings, the proportion
reaches approximately 1 in 5 \citep{Francis2024}. By 2040, CKD is projected to
rank fifth among leading causes of years of life lost globally
\citep{Francis2024}. Those numbers create genuine pressure on health systems to
identify high-risk patients early, before irreversible loss of kidney function
closes off the best treatment options.

Machine learning has been proposed as a practical answer to this challenge.
Models trained on electronic health records, biomarker panels, and demographic
variables have reported AUROC values above 0.95 in national cohort studies
\citep{Krishnamurthy2021, Bai2022, Li2025, Sabanayagam2025}. Established risk
equations like the Kidney Failure Risk Equation have been validated across
populations in North America, the United Kingdom, and Latin America, confirming
algorithmic CKD risk prediction is technically achievable
\citep{Tangri2016, Major2019, BravoZuniga2025}. The field has not struggled to
build models. The struggle is with what happens after the model is built.

Discrimination metrics like AUROC measure whether a model ranks patients
correctly relative to each other. They say nothing about whether the assigned
probability scores are trustworthy in absolute terms. A model posting AUROC 0.97
assigns a 65\% risk score to patients whose true event rate sits near 20\%, and
clinicians making treatment decisions from those numbers are working from
miscalibrated information. Van Calster and colleagues identified calibration as
the Achilles heel of predictive analytics, noting poor calibration regularly
persists even when discrimination appears strong \citep{VanCalster2019}. A
systematic review of CKD risk models by Echouffo-Tcheugui and Kengne found
calibration is assessed less commonly than discrimination across the published
literature; of all the models reviewed, only eight for CKD occurrence and five
for CKD progression had been externally validated for calibration
\citep{EchouffoTcheugui2012}. The models exist. The evidence for trusting their
probability outputs largely does not.

The problem runs deeper than calibration alone. Campagner and colleagues
reviewed machine learning studies in healthcare and found fewer than 4\% address
uncertainty quantification explicitly \citep{Campagner2025}. A model outputting
a probability without any indication of how much to trust the number puts
clinicians in a difficult position. Banerji and colleagues put this plainly:
clinical AI tools must communicate predictive uncertainty at the level of the
individual patient, not purely in aggregate performance statistics
\citep{Banerji2023}. A predicted CKD risk of 78\% warrants a different clinical
response when the model's uncertainty is narrow than when high uncertainty
renders the output practically unreliable.

A 2023 systematic review commissioned to support CDC prevention guidelines
reached a pointed conclusion: CKD risk prediction models need to be better
calibrated and externally validated before incorporation into clinical guidelines
\citep{GonzalezRocha2023}. No published study in the CKD literature has
operationalized all three of these demands together. Current guidance for clinical
prediction model evaluation calls for joint assessment of discrimination,
calibration, fairness, and generalizability before deployment consideration
\citep{Collins2024EvalPart1}, and the design of external validation studies
requires attention to population comparability and feature completeness
\citep{Riley2024EvalPart2}. No existing work has jointly
evaluated calibration across multiple post-hoc correction methods, quantified
uncertainty through a coverage-guaranteed conformal framework, and assessed
deployment readiness through a structured multi-criterion checklist, all on the
same model suite and across an independent external cohort.

This study addresses the gap. Using the UCI CKD dataset for model development
and MIMIC-IV as an external validation cohort, we trained five classifiers
spanning the range commonly used in clinical prediction: logistic regression,
random forest, gradient boosting via XGBoost, a support vector machine with
Platt-scaled probabilities, and Gaussian naive Bayes. Each model was evaluated
across three dimensions: calibration before and after post-hoc recalibration
using Platt scaling and isotonic regression; predictive uncertainty through split
conformal prediction with a formal 90\% marginal coverage guarantee; and a
structured eight-criterion deployment readiness framework grounded in current
reporting standards including TRIPOD+AI \citep{Collins2024TRIPOD}.

The study has three objectives:
\begin{enumerate}
    \item Quantify pre- and post-calibration error for five CKD classifiers on
          both the internal UCI test set and the external MIMIC-IV cohort.
    \item Apply split conformal prediction to generate prediction sets with a
          90\% coverage guarantee and determine whether the guarantee holds on
          external data.
    \item Score each model against an eight-criterion deployment readiness
          checklist and identify which, if any, meet the threshold for
          responsible clinical use.
\end{enumerate}

%% file: sections/methods.tex
\subsection{Datasets}

Two datasets were used. The UCI CKD dataset served as the primary training and internal validation source~\cite{UCIDataset}. It contains 400 patient records collected from a hospital in Vellore, India, each described by 24 clinical and laboratory features alongside a binary CKD label. Of the 400 patients, 250 (62.5\%) carry a positive CKD diagnosis. Mean patient age was 51.6 years (SD 17.0). Features include continuous measurements such as serum creatinine, blood urea, hemoglobin, sodium, potassium, and packed cell volume, plus categorical variables for comorbidities (hypertension, diabetes, coronary artery disease) and urinary findings (red blood cell morphology, pus cells, bacteria).

The MIMIC-IV Clinical Database Demo (version 2.2) provided a distributional stress-test cohort~\cite{Johnson2023}. This is a publicly available, open-access subset of MIMIC-IV released by PhysioNet specifically for pipeline development and workshop use; it contains 100 de-identified patients from Beth Israel Deaconess Medical Center in Boston and requires no credentialing. It is not a formally designed external validation set, and its use here is deliberately framed as a stress-test: the goal is to evaluate how each model behaves when applied to a population with different prevalence, missing features, and clinical context than the training data.

Patients were included if serum creatinine was available on their first hospital admission, yielding 97 patients. CKD labeling used the CKD-EPI 2021 race-free equation: eGFR below 60\,mL/min/1.73\,m\textsuperscript{2} defined CKD-positive status. That threshold produced 23 CKD cases (23.7\%) and 74 controls. Mean age in the demo cohort was 61.7 years (SD 16.3). The dataset is fully de-identified; no human subjects review was required.

\subsection{Preprocessing and Feature Harmonization}

UCI preprocessing started with whitespace stripping across all categorical columns. Categorical features were mapped to binary 0/1 values. The target label was encoded as CKD\,=\,1, notCKD\,=\,0. Missing values in continuous features were replaced with the column median; categorical missing values were replaced with the column mode. Zero missing values remained after imputation across all 400 records.

MIMIC harmonization used statistics from the UCI training fold only, to prevent any leakage from the external data into the model pipeline. Seven features present in the UCI schema are not routinely recorded in MIMIC: urine-specific gravity, urine sugar, pus cells (categorical), pus cell clumps, bacteria, appetite, and pedal edema. Each was filled with the UCI training-set median or mode as appropriate. Blood pressure came from ICU chartevents (item IDs 220179 and 220050, systolic range 60--250\,mmHg); where that source was missing, the MIMIC outpatient medical record table provided a fallback. That two-source approach resolved blood pressure for 94 of 97 patients. Laboratory values including serum creatinine, blood urea nitrogen, hemoglobin, albumin, potassium, sodium, glucose, hematocrit, WBC, and RBC were extracted from the MIMIC lab events table and averaged across each patient's first admission. Comorbidity flags for hypertension, diabetes, and coronary artery disease came from ICD-10 codes (I10, E11.x, I25.x). Anemia was defined as hemoglobin below 12\,g/dL in females and below 13.5\,g/dL in males.

\subsection{Model Suite and Training}

Records were split into training (70\%, $n$\,=\,279), validation (15\%, $n$\,=\,60), and test (15\%, $n$\,=\,61) subsets using stratified random sampling (random\_state\,=\,42). The MIMIC demo cohort was held back entirely as a stress-test set, with no records used during any training, validation, or calibration step.

Five classifier families were selected to span the calibration behaviors seen in clinical ML: logistic regression (L2 regularization), random forest (ensemble, known for overconfident probabilities), XGBoost (gradient boosting, strong discrimination but typically miscalibrated), support vector machine (\texttt{probability=True}), and Gaussian naive Bayes~\cite{NiculescuMizil2005}. Hyperparameter tuning used five-fold stratified cross-validation on the training fold, optimizing AUROC. Logistic regression $C$ was tuned over \{0.001, 0.01, 0.1, 1, 10, 100\}. RF, XGB, and SVM used randomized search with up to 30 iterations over defined grids (full grids in Supplementary S1). Fitted models were saved with joblib. Software: Python 3.13, scikit-learn, XGBoost, MAPIE 1.3, netcal, pandas, numpy, matplotlib, joblib; full version pins in requirements.txt.

\subsection{Calibration Evaluation}

Pre-calibration metrics were computed on the UCI validation set for each model: Expected Calibration Error (ECE, 10 equal-width bins)~\cite{Guo2017}, Maximum Calibration Error (MCE), Brier Score, and Brier Skill Score relative to a naive prevalence baseline. ECE and MCE used the netcal library. Reliability diagrams used \texttt{CalibrationDisplay} from scikit-learn.

Two post-hoc recalibration methods were fitted on the validation set. Platt scaling fits a logistic regression layer on the base model's raw scores~\cite{Platt1999} (\texttt{CalibratedClassifierCV}, \texttt{method='sigmoid'}, \texttt{cv='prefit'}). Isotonic regression fits a piecewise-constant monotone function~\cite{Zadrozny2002} (\texttt{method='isotonic'}). \texttt{FrozenEstimator} prevented any refitting of the base model in both cases. The test set was not used at any point during calibration fitting.

Post-calibration metrics were computed on the UCI test set for the base, Platt-scaled, and isotonic-scaled variants. For external validation, the best variant per model (lowest ECE on the UCI test set) was applied to MIMIC. Calibration drift was MIMIC ECE minus UCI ECE for that variant.

\subsection{Uncertainty Quantification}

Uncertainty was quantified through split conformal prediction using the MAPIE library (\texttt{SplitConformalClassifier}, version 1.3)~\cite{Taquet2022,AngelopoulosBates2022}. Each base model's conformal predictor was fitted on the UCI validation set ($n$\,=\,60) using the least ambiguous class (LAC) conformity score: one minus the predicted probability of the most likely class. Target confidence was 0.90 ($\alpha$\,=\,0.10), so prediction sets should contain the true label for at least 90\% of test cases.

Three metrics were computed on both the UCI test set and the MIMIC cohort: empirical coverage rate, average prediction set size, and singleton rate (the share of cases receiving exactly one class label). Coverage drift was UCI coverage minus MIMIC coverage.

\subsection{Deployment Readiness Framework}

Eight criteria were defined before analysis began, drawing on reporting standards for early-stage clinical AI evaluation~\cite{Vasey2022,Collins2024TRIPOD}. Each was scored PASS (threshold met, 2 points), MARGINAL (within 20\% of threshold, 1 point), or FAIL (0 points), for a maximum total of 16 points.

\begin{enumerate}
  \item Discrimination adequacy: AUROC $\geq 0.85$ on the external cohort.
  \item Calibration adequacy: ECE $\leq 0.10$ on the external cohort.
  \item Calibration stability: absolute calibration drift $\leq 0.05$.
  \item Uncertainty coverage: conformal coverage $\geq 0.90$ on the external cohort.
  \item Coverage stability: absolute coverage drift $\leq 0.05$.
  \item Prediction interpretability: singleton rate $\geq 0.70$ on the external cohort.
  \item Subgroup calibration equity: maximum ECE gap across subgroups $\leq 0.05$.
  \item Transparency: full code and pipeline publicly available (automatic PASS).
\end{enumerate}

Subgroup analysis stratified the MIMIC cohort by age (below 65 vs.\ 65 and above), diabetes status, and hypertension status. Groups with fewer than 10 patients were excluded from ECE computation because bin-level estimates become unreliable at that scale.

\subsection{Statistical Analysis}

All metrics are point estimates on held-out test sets. Bootstrap confidence intervals (95\%, 1000 resamples, random\_state\,=\,42) were computed for AUROC and ECE on both cohorts. No hypothesis tests were run; the study is descriptive and evaluative.

%% file: sections/results.tex
\subsection{Cohort Characteristics}

The UCI dataset included 400 patients, mean age 51.6 years (SD 17.0), CKD prevalence 62.5\% ($n$\,=\,250). Fourteen continuous and 10 binary categorical features were present. Preprocessing left zero missing values. The 70/15/15 split produced 279 training, 60 validation, and 61 test patients, each fold within one percentage point of the overall CKD prevalence.

The MIMIC-IV demo stress-test cohort had 97 patients, mean age 61.7 years (SD 16.3). CKD prevalence was 23.7\% ($n$\,=\,23), substantially lower than the UCI training population (62.5\%), and reflecting a general hospital population rather than a nephrology referral center. Seven of the 24 model features were entirely absent from the demo and were imputed using UCI training-set statistics, meaning those columns carry no information specific to individual MIMIC patients. Blood pressure was recovered for 94 of 97 patients through ICU chartevents and outpatient records.

\subsection{Baseline Discrimination}

All five models reached AUROC 1.00 on the UCI test set (Table~\ref{tab:t1}). The dataset offers essentially no discrimination challenge once models are fitted. RF and SVM posted perfect F1, accuracy, sensitivity, and specificity. LR lagged on accuracy (0.77) and specificity (0.39), a pattern that points to a miscalibrated intercept pushing most probabilities above the decision threshold. XGB and NB achieved F1 of 0.987. Cross-validation AUROC during tuning was 0.999--1.000 across the board. The UCI benchmark appears saturated.

% ── Table 1: Baseline Discrimination ──────────────────────────────────────────
\begin{table}[ht]
\centering
\caption{Baseline discrimination on the UCI test set ($n$\,=\,61). All five models achieve AUROC 1.00. LR specificity of 0.391 reflects a miscalibrated decision boundary rather than poor discrimination.}
\label{tab:t1}
\begin{tabular}{lcccccc}
\toprule
Model & AUROC & AUPRC & F1 & Accuracy & Sensitivity & Specificity \\
\midrule
LR  & 1.000 & 1.000 & 0.844 & 0.770 & 1.000 & 0.391 \\
RF  & 1.000 & 1.000 & 1.000 & 1.000 & 1.000 & 1.000 \\
XGB & 1.000 & 1.000 & 0.987 & 0.984 & 0.974 & 1.000 \\
SVM & 1.000 & 1.000 & 1.000 & 1.000 & 1.000 & 1.000 \\
NB  & 1.000 & 1.000 & 0.987 & 0.984 & 0.974 & 1.000 \\
\bottomrule
\end{tabular}
\end{table}

\subsection{Pre-Calibration Assessment}

\begin{figure}[H]
\centering
\begin{subfigure}[b]{0.70\textwidth}
  \includegraphics[width=\textwidth]{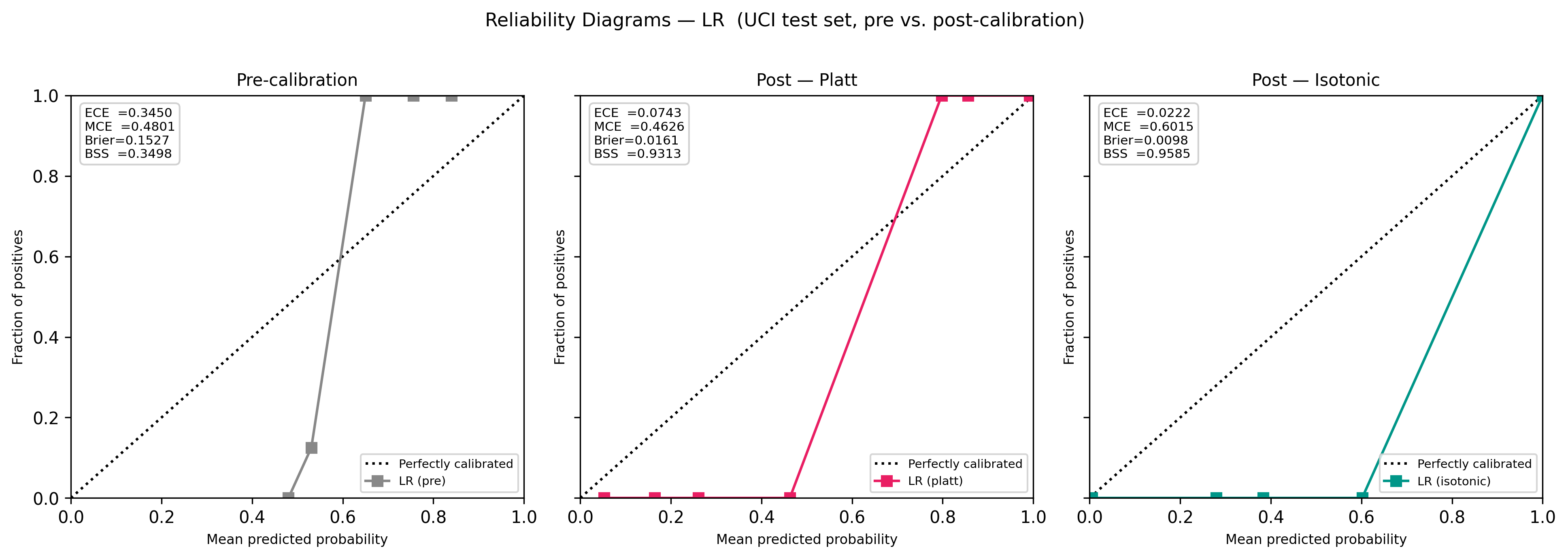}
  \caption{LR}
\end{subfigure}
\\[0.6em]
\begin{subfigure}[b]{0.70\textwidth}
  \includegraphics[width=\textwidth]{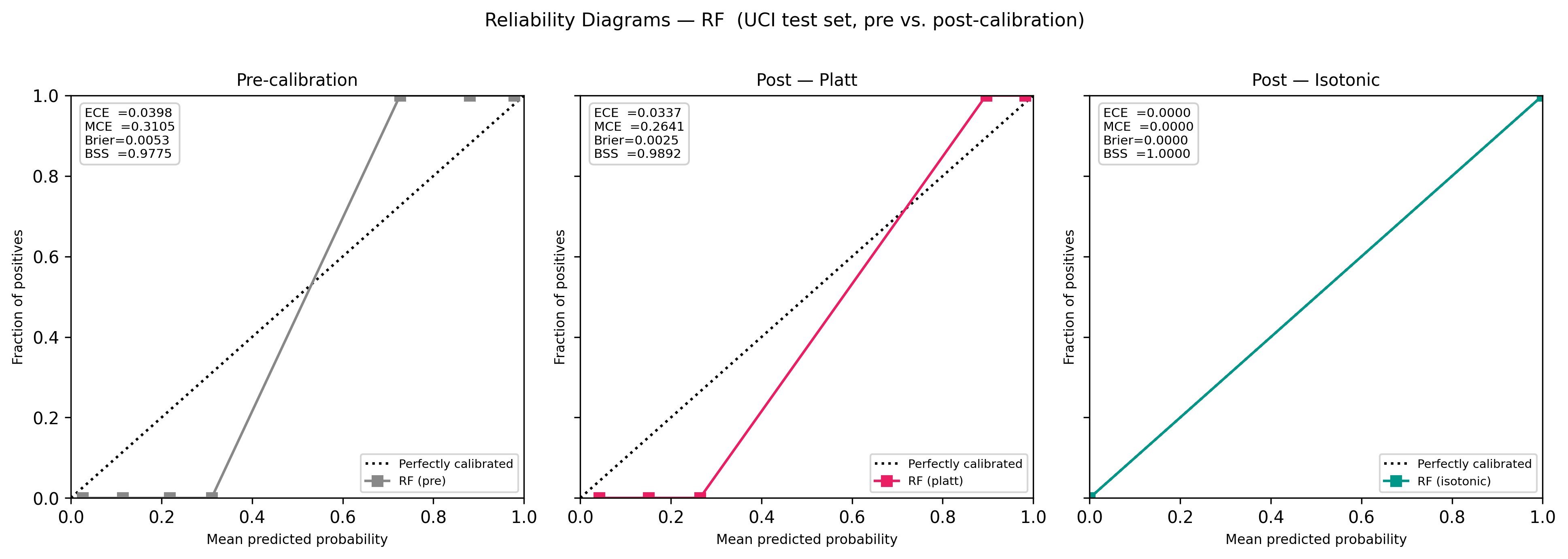}
  \caption{RF}
\end{subfigure}
\\[0.6em]
\begin{subfigure}[b]{0.70\textwidth}
  \includegraphics[width=\textwidth]{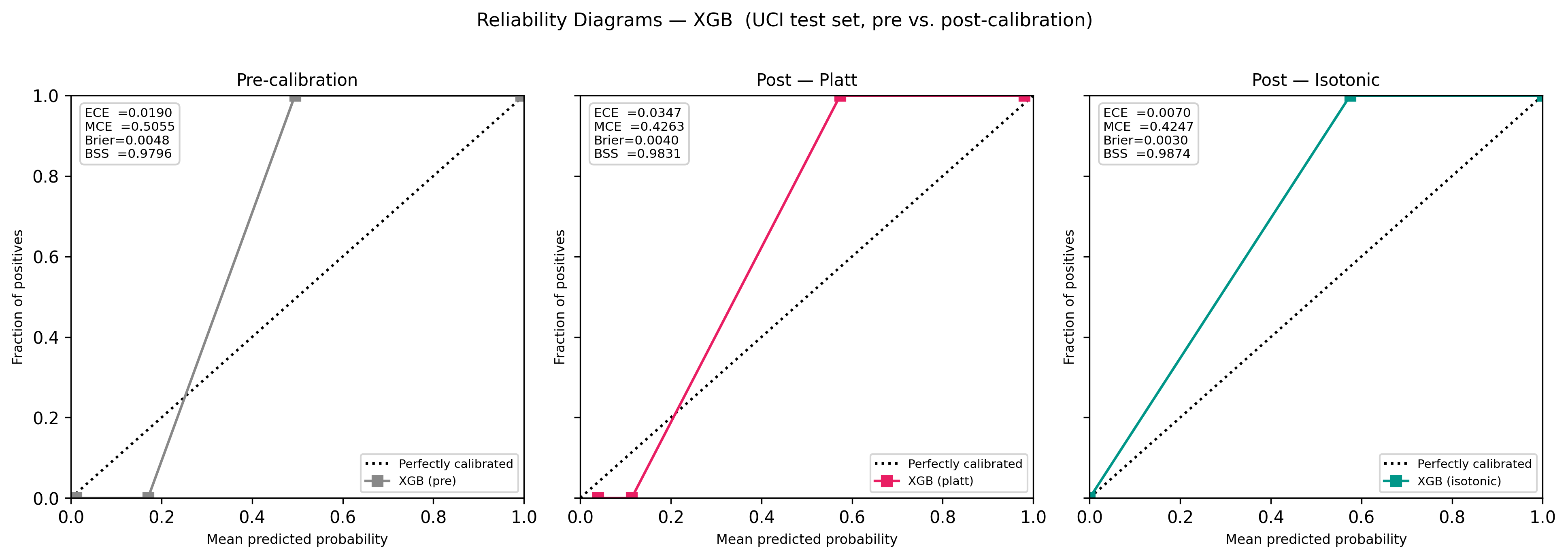}
  \caption{XGB}
\end{subfigure}
\caption{Reliability diagrams for LR, RF, and XGB before and after post-hoc calibration on the UCI test set. Each panel shows the uncalibrated model (grey), Platt scaling (pink), and isotonic regression (teal). The diagonal is perfect calibration. LR shows the largest pre-calibration deviation. Isotonic regression achieves the largest ECE reductions. (Continued in Figure~\ref{fig:f1b}.)}
\label{fig:f1a}
\end{figure}

\begin{figure}[H]
\centering
\begin{subfigure}[b]{0.70\textwidth}
  \includegraphics[width=\textwidth]{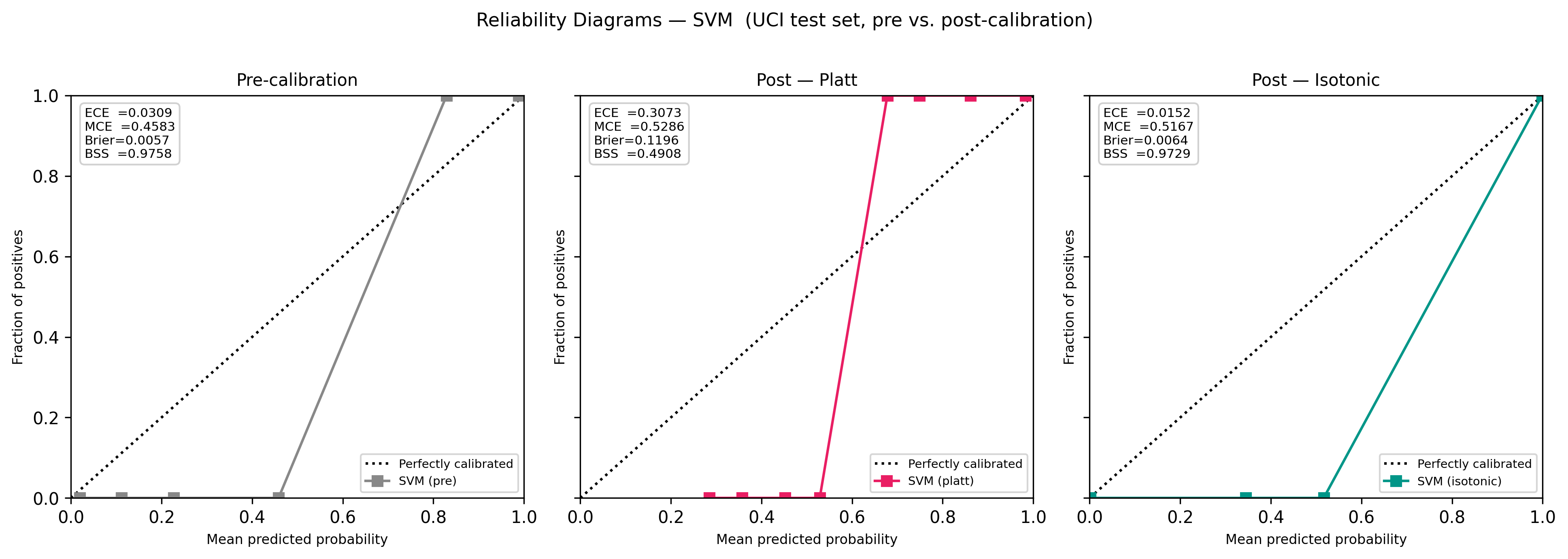}
  \caption{SVM}
\end{subfigure}
\\[0.6em]
\begin{subfigure}[b]{0.70\textwidth}
  \includegraphics[width=\textwidth]{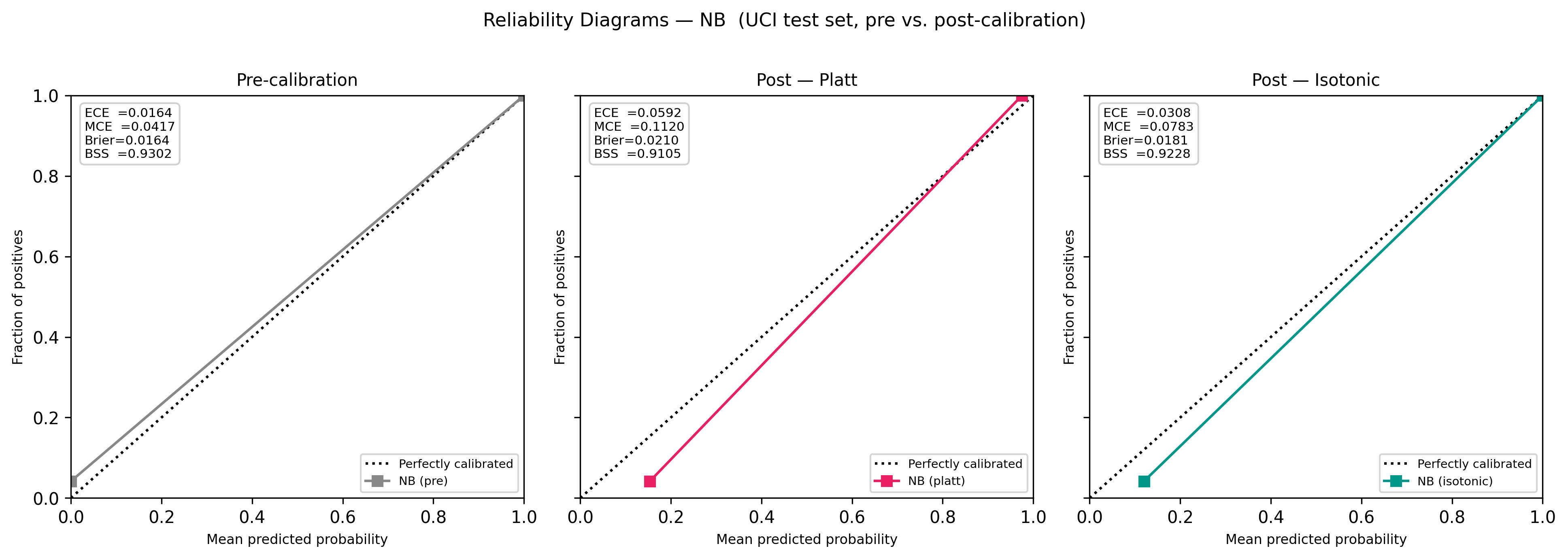}
  \caption{NB}
\end{subfigure}
\caption{Reliability diagrams for SVM and NB before and after post-hoc calibration on the UCI test set (continued from Figure~\ref{fig:f1a}). Platt scaling raised SVM calibration error rather than reducing it. NB's uncalibrated variant outperformed both post-hoc variants.}
\label{fig:f1b}
\end{figure}

Before any post-hoc correction, calibration varied widely. LR was the weakest: ECE 0.263, MCE 0.483, Brier Score 0.164, Brier Skill Score 0.295. XGB was the strongest, with ECE 0.031, Brier Score 0.005, and a Brier Skill Score of 0.977. RF, SVM, and NB fell in between with ECE values of 0.053, 0.042, and 0.050. Reliability diagrams confirmed LR's systematic bias: underestimation at low predicted values, overestimation at the high end. XGB and SVM stayed close to the diagonal throughout (Figures~\ref{fig:f1a} and~\ref{fig:f1b}).

% ── Figure 1a: Reliability Diagrams (LR, RF, XGB) ────────────────────────────

% ── Figure 1b: Reliability Diagrams (SVM, NB) ────────────────────────────────

\subsection{Post-Calibration Results}

Isotonic regression cut ECE across the full model suite (Table~\ref{tab:t2}). LR dropped from 0.345 to 0.022 (95\% CI: 0.002, 0.050), a reduction of 0.323. RF reached ECE 0.000. XGB went from 0.019 to 0.007 (95\% CI: 0.000, 0.021). SVM improved from 0.031 to 0.015 (95\% CI: 0.000, 0.037). For NB, the uncalibrated model (ECE 0.016, 95\% CI: 0.000, 0.049) outperformed both post-hoc variants, so it was selected as NB's best model. Platt scaling produced inconsistent results: it helped LR but raised SVM's ECE from 0.031 to 0.307, likely because the SVM's decision scores and true probabilities do not have a monotone relationship on this dataset.

\subsection{Distributional Stress-Test on MIMIC-IV Demo}

\begin{figure}[H]
\centering
% Row 1
\begin{subfigure}[b]{0.27\textwidth}
  \includegraphics[width=\textwidth]{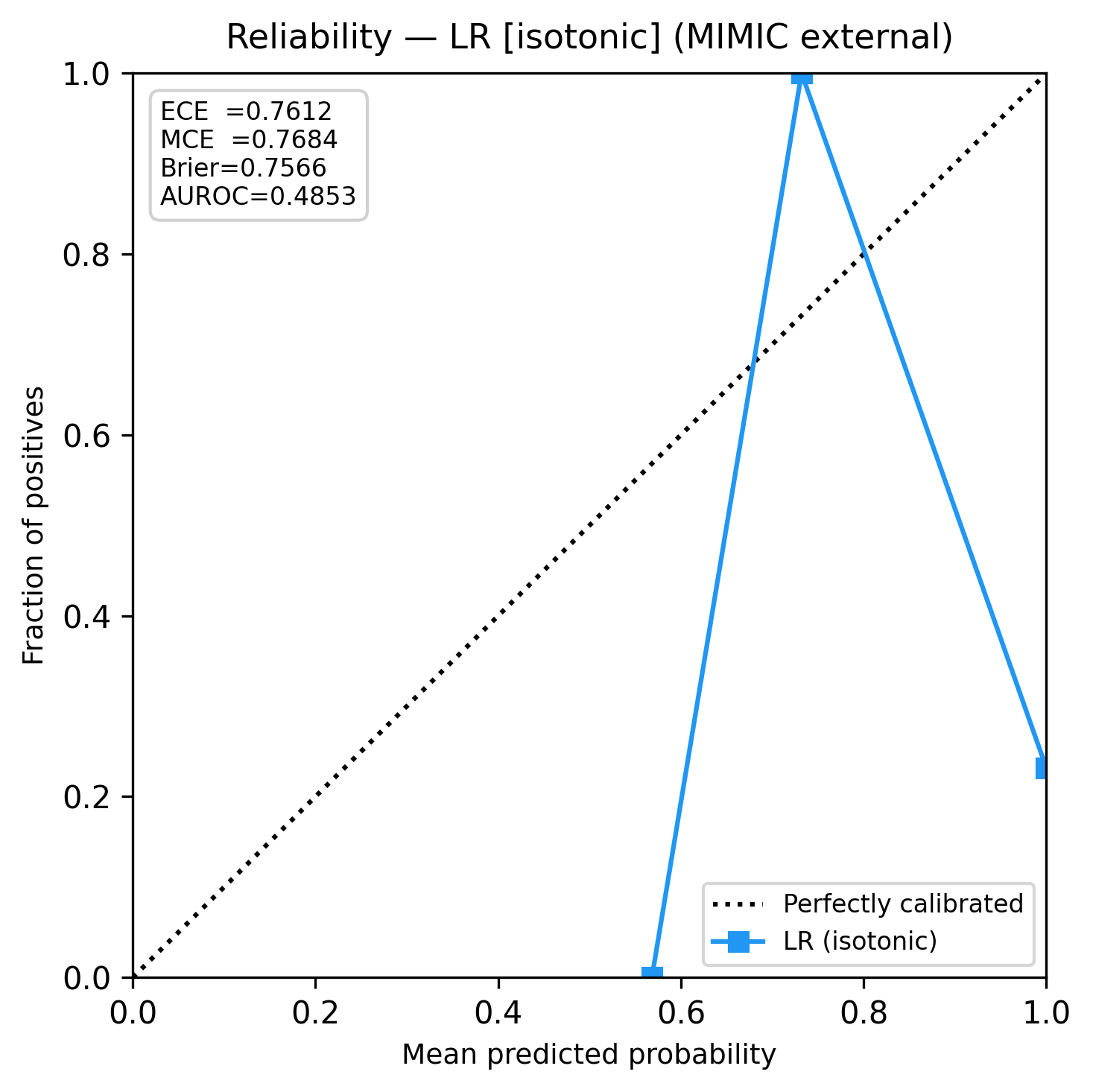}
  \caption{LR}
\end{subfigure}
\hfill
\begin{subfigure}[b]{0.27\textwidth}
  \includegraphics[width=\textwidth]{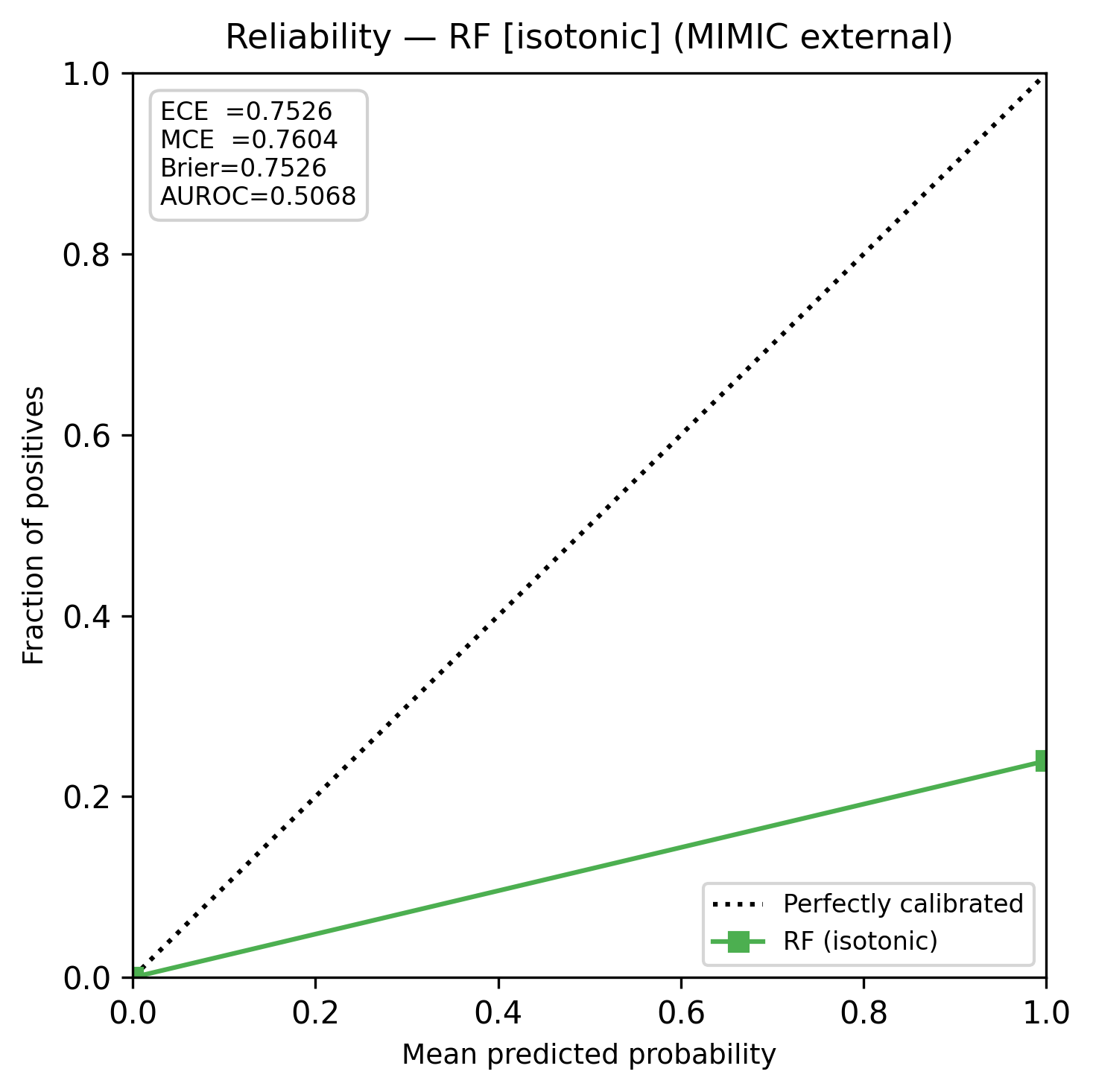}
  \caption{RF}
\end{subfigure}
\hfill
\begin{subfigure}[b]{0.27\textwidth}
  \includegraphics[width=\textwidth]{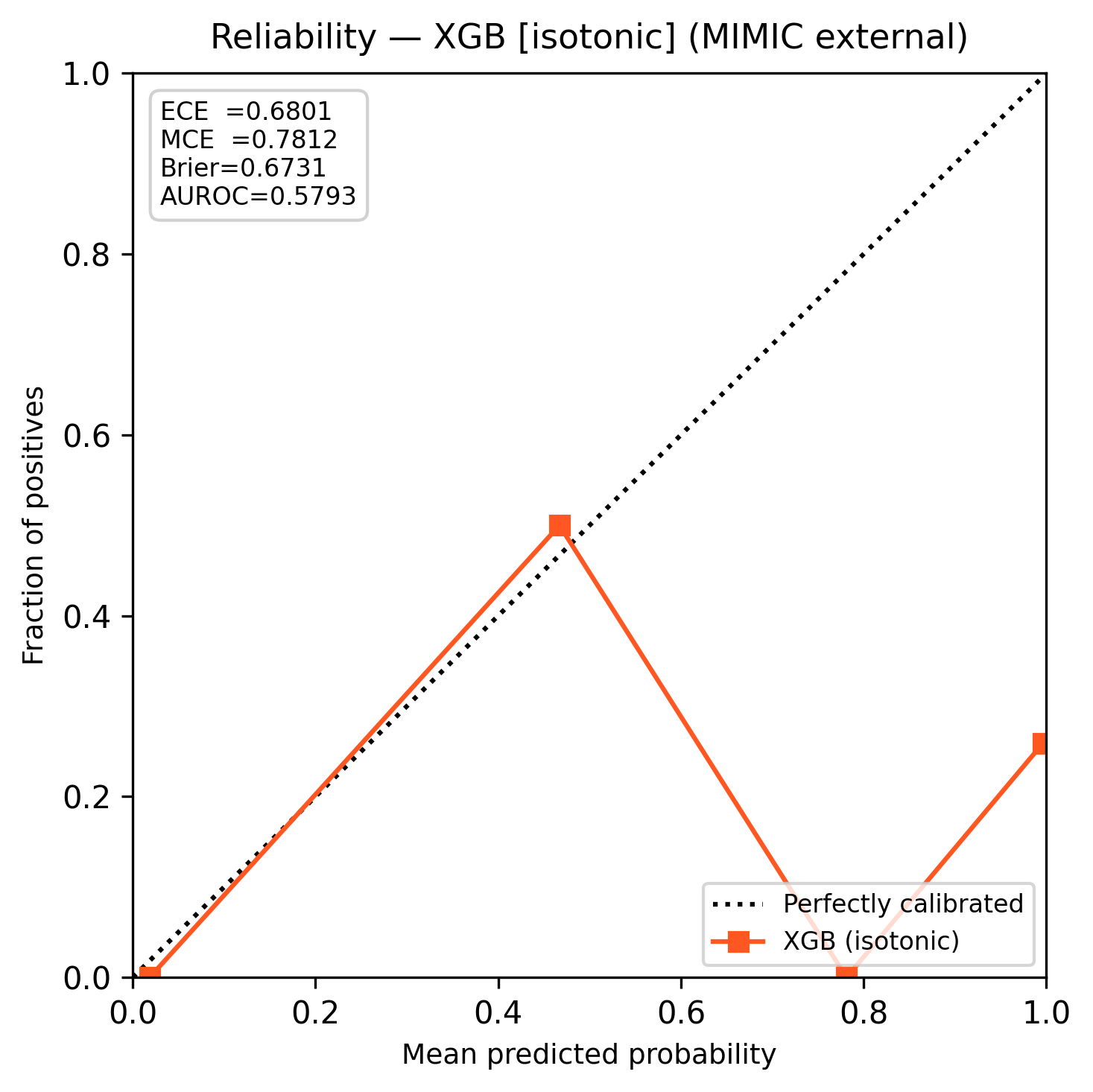}
  \caption{XGB}
\end{subfigure}
\\[0.6em]
% Row 2
\begin{subfigure}[b]{0.27\textwidth}
  \includegraphics[width=\textwidth]{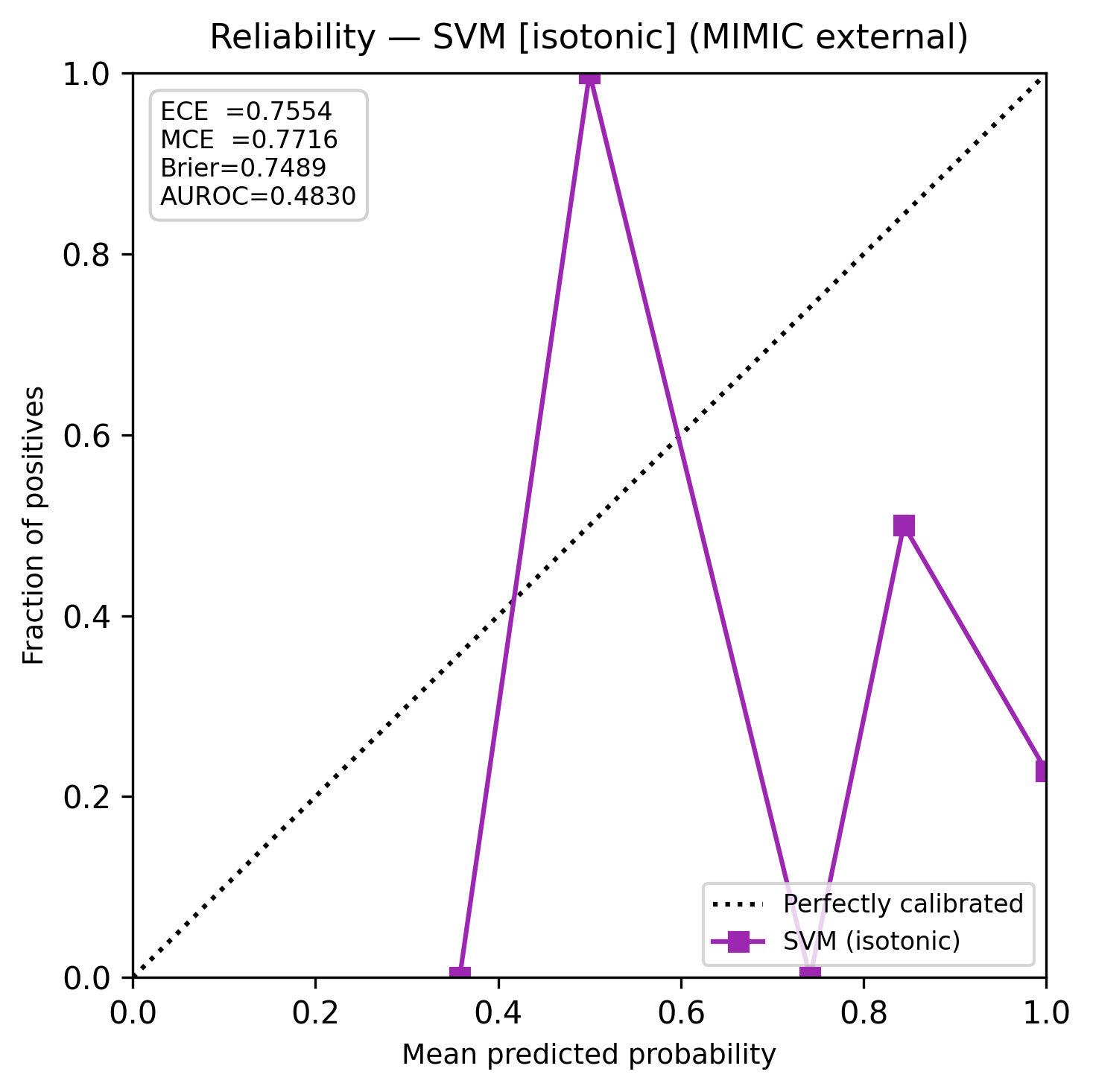}
  \caption{SVM}
\end{subfigure}
\hfill
\begin{subfigure}[b]{0.27\textwidth}
  \includegraphics[width=\textwidth]{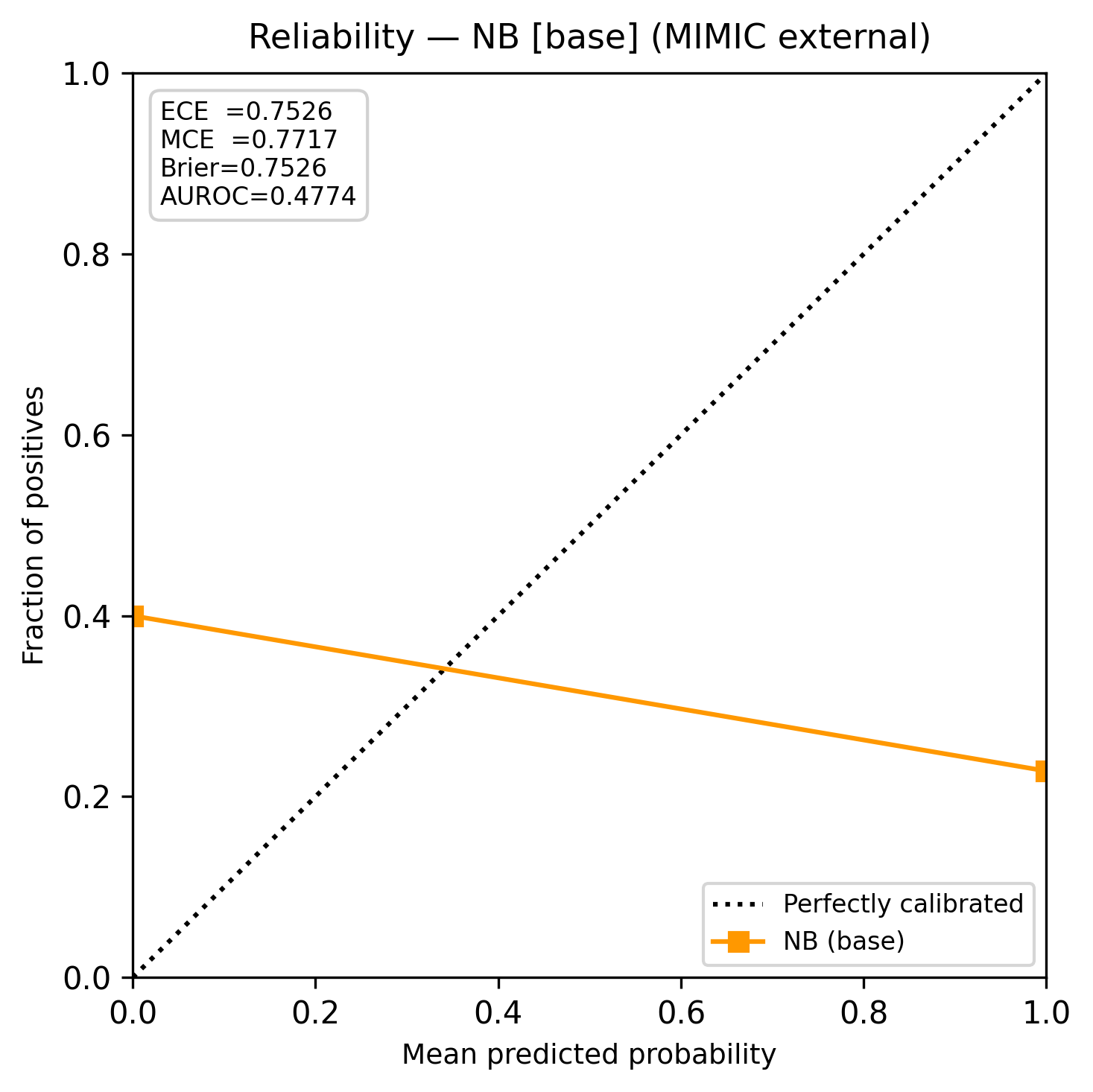}
  \caption{NB}
\end{subfigure}
\caption{Reliability diagrams for the best-calibrated variant of each classifier on the MIMIC-IV demo stress-test cohort. All curves fall below the diagonal, consistent with models trained on a 62.5\% CKD prevalence population being applied to a 23.7\% prevalence cohort with seven imputed features. ECE ranged from 0.680 (XGB) to 0.761 (LR). Results demonstrate distributional failure, not clinical deployment performance.}
\label{fig:f2}
\end{figure}

Applying the best-calibrated variant of each model to the MIMIC demo cohort produced a sharp deterioration across every metric (Table~\ref{tab:t2}, Figure~\ref{fig:f2}). This outcome is expected given the two distributional differences: a 39-percentage-point prevalence gap and seven features imputed from UCI statistics rather than measured from individual patients. AUROC fell to 0.485 (LR), 0.507 (RF), 0.579 (XGB), 0.483 (SVM), and 0.477 (NB), all near or below chance. ECE reached 0.761 for LR (95\% CI: 0.673, 0.844), 0.753 for RF (95\% CI: 0.660, 0.835), 0.680 for XGB (95\% CI: 0.594, 0.777), 0.755 for SVM (95\% CI: 0.667, 0.837), and 0.753 for NB (95\% CI: 0.660, 0.835). Calibration drift ranged from 0.673 (XGB) to 0.753 (RF), far above the 0.05 threshold. XGB was the least poor performer on both AUROC and ECE, a pattern consistent with gradient boosting learning slightly more transferable representations under regularization.

% ── Table 2: Calibration Summary ──────────────────────────────────────────────
\begin{table}[ht]
\centering
\caption{Calibration summary for all five classifiers. Best variant selected by lowest ECE on the UCI test set. Bootstrap 95\% CIs in brackets. Calibration drift = MIMIC demo ECE $-$ UCI ECE. MIMIC results reflect distributional shift, not formal external validation.}
\label{tab:t2}
\begin{tabular}{llcccc}
\toprule
 & & \multicolumn{2}{c}{UCI Test Set} & \multicolumn{2}{c}{MIMIC-IV} \\
\cmidrule(lr){3-4} \cmidrule(lr){5-6}
Model & Variant & ECE [95\% CI] & AUROC & ECE [95\% CI] & AUROC \\
\midrule
LR  & Isotonic & 0.022 [0.002, 0.050] & 1.000 & 0.761 [0.673, 0.844] & 0.485 \\
RF  & Isotonic & 0.000 [0.000, 0.000] & 1.000 & 0.753 [0.660, 0.835] & 0.507 \\
XGB & Isotonic & 0.007 [0.000, 0.021] & 1.000 & 0.680 [0.594, 0.777] & 0.579 \\
SVM & Isotonic & 0.015 [0.000, 0.037] & 1.000 & 0.755 [0.667, 0.837] & 0.483 \\
NB  & Base     & 0.016 [0.000, 0.049] & 1.000 & 0.753 [0.660, 0.835] & 0.477 \\
\midrule
\multicolumn{6}{l}{\footnotesize Calibration drift (MIMIC ECE $-$ UCI ECE): LR 0.739, RF 0.753, XGB 0.673, SVM 0.740, NB 0.736.}\\
\bottomrule
\end{tabular}
\end{table}

% ── Figure 2: MIMIC Reliability Diagrams ──────────────────────────────────────

\subsection{Conformal Prediction Coverage}

\begin{figure}[H]
\centering
\includegraphics[width=0.90\textwidth]{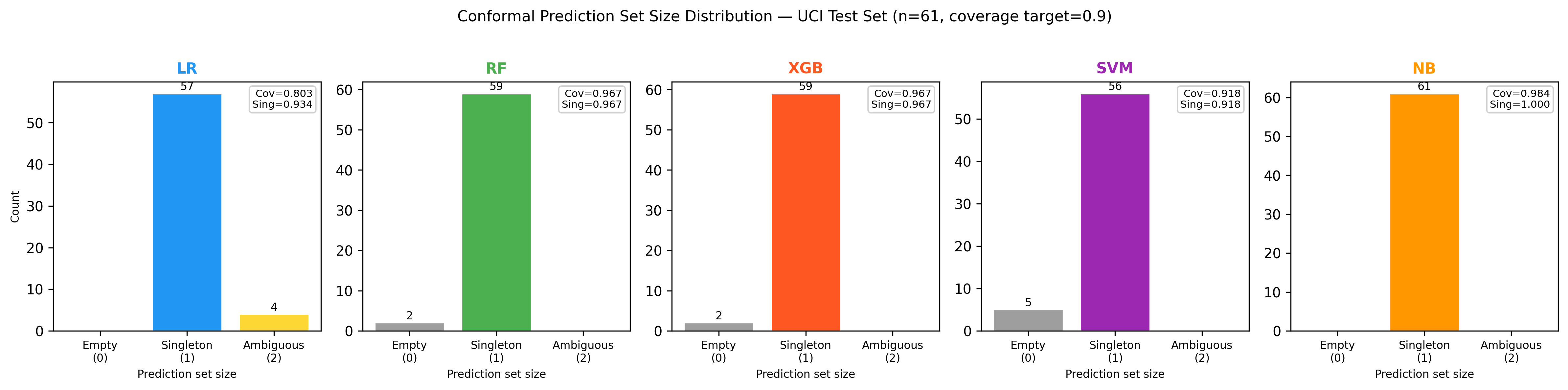}
\caption{Prediction set size distribution (UCI test set). Bar charts of set size (0\,=\,empty, 1\,=\,singleton, 2\,=\,ambiguous) per model. NB and RF produce the fewest ambiguous predictions. Coverage target is 0.90. (Individual-level display in Figure~\ref{fig:f3b}.)}
\label{fig:f3a}
\end{figure}

\begin{figure}[H]
\centering
\includegraphics[width=0.62\textwidth]{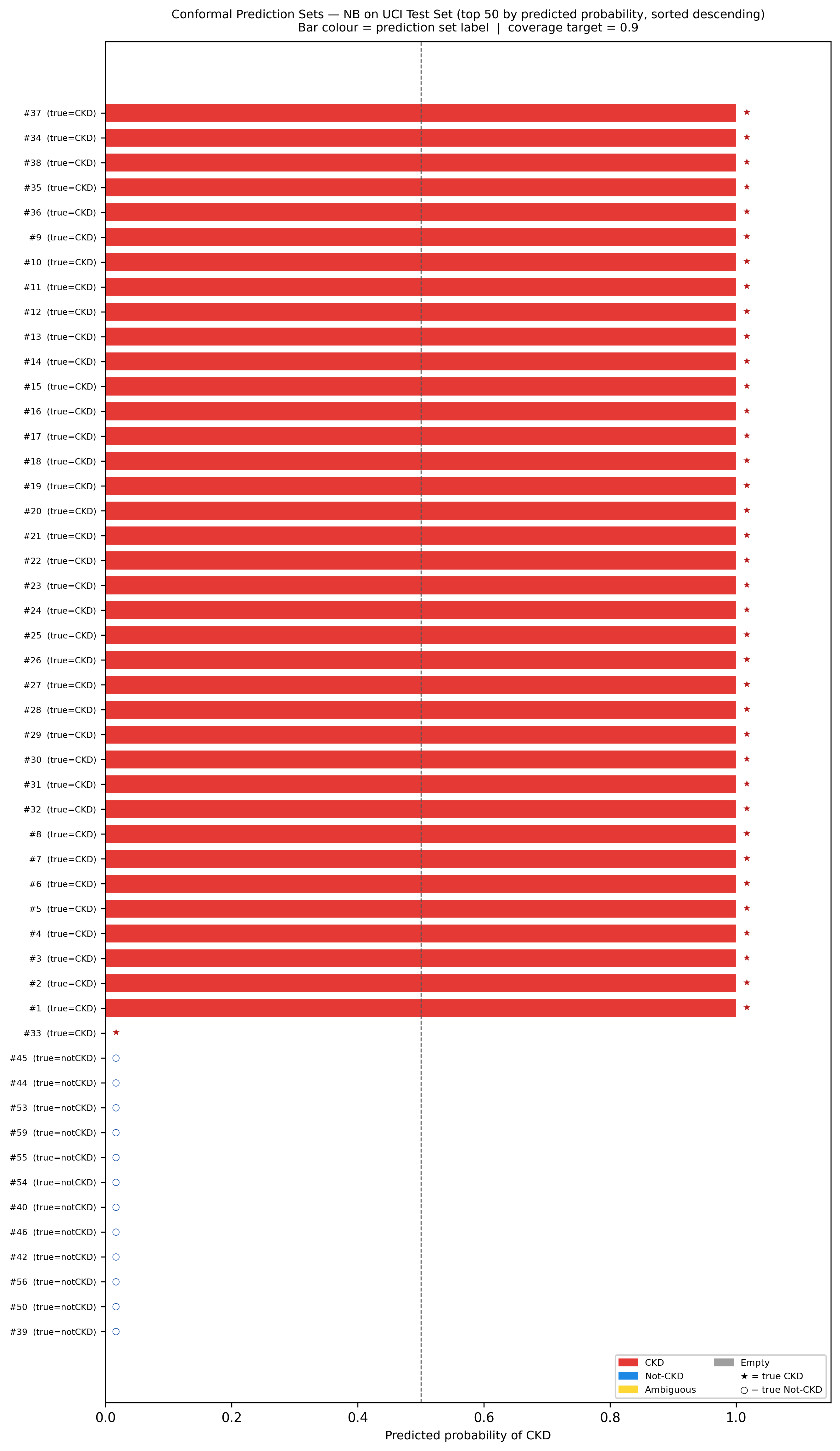}
\caption{Individual-level uncertainty display for NB (top 50 UCI test patients, sorted by descending predicted probability). Bar colour reflects prediction set membership: red\,=\,CKD only, blue\,=\,not-CKD only, yellow\,=\,ambiguous (both classes). Markers indicate true label. Coverage target is 0.90. (Set size distribution in Figure~\ref{fig:f3a}.)}
\label{fig:f3b}
\end{figure}

On the UCI test set, four of five models met the 90\% coverage target (Table~\ref{tab:t3}, Figures~\ref{fig:f3a} and~\ref{fig:f3b}). NB achieved 0.984 coverage with a singleton rate of 1.00. RF and XGB both reached 0.967 coverage and singleton rates of 0.967. SVM achieved 0.918. LR fell short at 0.803, consistent with its miscalibrated probabilities undermining the conformity scores.

Coverage collapsed on MIMIC. NB was highest at 0.247; the others ranged from 0.206 to 0.237. All five models fell below 0.30 against a 0.90 target. Coverage drift ranged from 0.566 (LR) to 0.761 (RF). The conformal guarantee holds only when calibration and test data come from the same distribution. These two cohorts are not exchangeable in that sense.

% ── Table 3: Uncertainty Summary ──────────────────────────────────────────────
\begin{table}[ht]
\centering
\caption{Conformal prediction results on the UCI test set and MIMIC-IV demo stress-test cohort (target coverage 0.90). Coverage drift = UCI coverage $-$ MIMIC coverage. Coverage collapse on MIMIC reflects distributional non-exchangeability, not a failure of the conformal method itself.}
\label{tab:t3}
\begin{tabular}{lcccccc}
\toprule
 & \multicolumn{3}{c}{UCI Test Set} & \multicolumn{3}{c}{MIMIC-IV} \\
\cmidrule(lr){2-4} \cmidrule(lr){5-7}
Model & Coverage & Avg Set & Singleton & Coverage & Avg Set & Singleton \\
\midrule
LR  & 0.803 & 1.066 & 0.934 & 0.237 & 1.000 & 1.000 \\
RF  & 0.967 & 0.967 & 0.967 & 0.206 & 0.804 & 0.804 \\
XGB & 0.967 & 0.967 & 0.967 & 0.227 & 0.474 & 0.474 \\
SVM & 0.918 & 0.918 & 0.918 & 0.216 & 0.938 & 0.938 \\
NB  & 0.984 & 1.000 & 1.000 & 0.247 & 0.990 & 0.990 \\
\midrule
\multicolumn{7}{l}{\footnotesize Coverage drift: LR 0.566, RF 0.761, XGB 0.740, SVM 0.701, NB 0.736.}\\
\bottomrule
\end{tabular}
\end{table}

% ── Figure 3a: Conformal Prediction Set Size Distribution ────────────────────

% ── Figure 3b: Individual-Level Uncertainty Display ───────────────────────────

\subsection{Deployment Readiness Scores}

\begin{figure}[H]
\centering
\includegraphics[width=\textwidth]{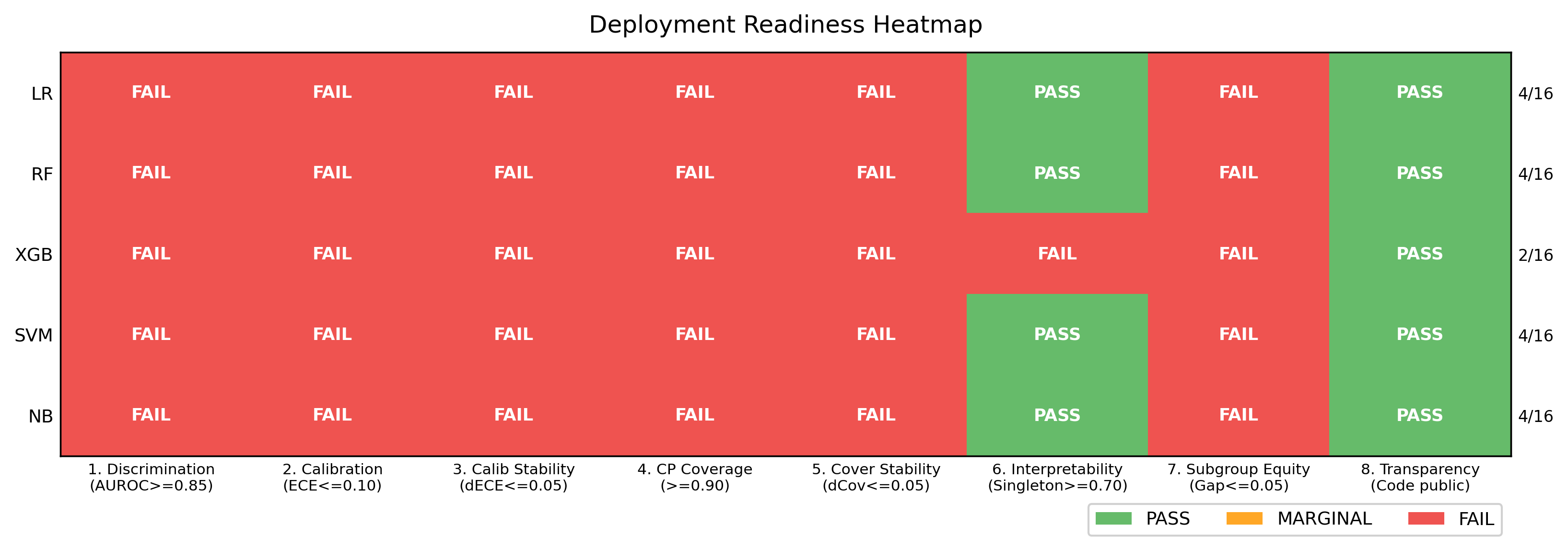}
\caption{Deployment readiness heatmap. Rows are models; columns are the eight criteria. Green\,=\,PASS, orange\,=\,MARGINAL, red\,=\,FAIL. Total scores out of 16 are shown on the right. All models score 2--4 out of 16. The only passing criteria are interpretability (C6) and transparency (C8). Criteria C1--C7 that reference the stress-test cohort all fail, illustrating that internal performance does not predict behaviour under distribution shift.}
\label{fig:f4}
\end{figure}

No model passed the checklist (Table~\ref{tab:t4}, Figure~\ref{fig:f4}). Scores ran from 2 out of 16 (XGB) to 4 out of 16 (LR, RF, SVM, NB). The only passing criteria were prediction interpretability on UCI (singleton rate above 0.70) and transparency. Every model failed discrimination on MIMIC, calibration adequacy on MIMIC, calibration stability, conformal coverage on MIMIC, coverage stability, and subgroup equity. Subgroup ECE gaps across age strata ranged from 0.148 to 0.209, far above the 0.05 equity threshold. XGB scored lowest (2/16) because its MIMIC singleton rate of 0.474 also failed the interpretability criterion; the other four models passed interpretability on UCI but not on MIMIC.

% ── Table 4: Deployment Checklist ─────────────────────────────────────────────
\begin{table}[ht]
\centering
\caption{Deployment readiness scores for all five classifiers across eight criteria. P\,=\,PASS (2 pts), M\,=\,MARGINAL (1 pt), F\,=\,FAIL (0 pts). Maximum score is 16.}
\label{tab:t4}
\begin{tabular}{lcccccccccc}
\toprule
Model & Variant & C1 & C2 & C3 & C4 & C5 & C6 & C7 & C8 & Total \\
\midrule
LR  & Isotonic & F & F & F & F & F & P & F & P & 4/16 \\
RF  & Isotonic & F & F & F & F & F & P & F & P & 4/16 \\
XGB & Isotonic & F & F & F & F & F & F & F & P & 2/16 \\
SVM & Isotonic & F & F & F & F & F & P & F & P & 4/16 \\
NB  & Base     & F & F & F & F & F & P & F & P & 4/16 \\
\bottomrule
\multicolumn{11}{l}{\footnotesize C1\,=\,Discrimination; C2\,=\,Calibration; C3\,=\,Calib.\ Stability; C4\,=\,CP Coverage;}\\
\multicolumn{11}{l}{\footnotesize C5\,=\,Coverage Stability; C6\,=\,Interpretability; C7\,=\,Subgroup Equity; C8\,=\,Transparency.}\\
\end{tabular}
\end{table}

% ── Figure 4: Deployment Heatmap ──────────────────────────────────────────────

%% file: sections/discussion.tex
\subsection{What the Distributional Stress-Test Reveals}

Every model in this study achieved AUROC 1.00 on the internal UCI test set. By the standard metrics that dominate published clinical ML literature, all five would be described as performing excellently. When the same models were applied to the MIMIC-IV demo cohort, AUROC fell to values indistinguishable from chance (0.48--0.58), ECE exceeded 0.68 for every model, and conformal coverage dropped from near-target to 0.21--0.25.

Before interpreting these numbers, the stress-test framing matters. The MIMIC demo is an open-access 100-patient subset released for pipeline development and workshop use, not a formally designed validation cohort. Its CKD prevalence (23.7\%) differs from the UCI training population (62.5\%) by 39 percentage points, and seven of 24 model features are entirely absent from the demo data. The purpose of including it is to illustrate what happens when models encounter a population that does not resemble their training data, not to estimate clinical deployment performance.

That said, the pattern is exactly what calibration theory predicts. Isotonic recalibration that achieved ECE 0.000 internally did not survive the shift. Echouffo-Tcheugui and Kengne found external calibration validation to be the exception rather than the norm in CKD prediction literature~\cite{EchouffoTcheugui2012}. Liou and colleagues documented a similar gap in a deployed malnutrition prediction model, finding calibration drift and subgroup bias that required post-deployment recalibration within a large healthcare system~\cite{Liou2024}. This study provides a concrete, reproducible illustration of why that gap matters: perfect internal metrics say nothing about behaviour under distribution shift.

\subsection{What Drove the External Failure}

Three factors contributed to the calibration collapse. The first is prevalence shift. A model trained and calibrated on a 62.5\% CKD population assigns probabilities near the training prevalence. The MIMIC cohort sits at 23.7\% CKD, so those probability estimates are systematically too high, producing calibration curves that fall above the diagonal at every predicted probability level. This is the most direct driver of high MIMIC ECE.

The second factor is feature missingness. Seven of the 24 features in the UCI schema are not routinely recorded in MIMIC. Urine-specific gravity, urine sugar, pus cells, pus cell clumps, bacteria, appetite, and pedal edema were all imputed using UCI training-set medians and modes. Those imputed values carry no information about individual MIMIC patients. From the model's perspective, seven features are effectively noise columns on the external cohort.

The third factor is dataset saturation. AUROC 1.00 on a 61-patient test set, with near-perfect cross-validation scores, signals that the UCI benchmark does not offer a realistic discrimination challenge. The learned decision boundaries are sharp enough to separate every patient in the training domain. Those boundaries do not generalize to a population with a different disease severity distribution and different measurement patterns.

XGB was the least poor performer externally (AUROC 0.579, ECE 0.680). That small margin appears to reflect gradient boosting's regularization limiting overfitting to UCI-specific patterns. Even so, the margin is not large enough to change the practical conclusion.

\subsection{Conformal Prediction as a Diagnostic Tool}

The conformal coverage collapse offers something calibration numbers alone do not: a direct, interpretable signal of distribution shift. Conformal prediction has been applied in clinical settings precisely because it provides this kind of interpretable coverage guarantee, including recent applications to individual-level diagnostic uncertainty in chronic disease~\cite{Vazquez2022,Sreenivasan2025}. The theoretical guarantee of split conformal prediction holds only when calibration and test data come from the same distribution~\cite{AngelopoulosBates2022}. Coverage of 0.22--0.25 on MIMIC, against a 0.90 target, is a quantitative statement that the MIMIC population is not exchangeable with the UCI validation set. A system operator who sees conformal coverage fall from 0.97 to 0.22 gets an immediate signal that the model is operating outside its valid domain.

That is the practical value of including conformal prediction in a deployment framework, separate from its theoretical properties. It creates a coverage audit trail. High MIMIC singleton rates for LR (1.00) and NB (0.99) might appear to suggest interpretable outputs. They do not: when coverage is only 0.24, a singleton prediction is a confident wrong answer for roughly three quarters of patients. Singleton rate without coverage context is not a useful interpretability metric.

\subsection{Comparison to Published CKD Models}

The eight externally-validated CKD models identified by Echouffo-Tcheugui and Kengne reported a range of calibration outcomes, but assessment was typically informal, often through visual calibration plots rather than ECE or Brier Score computation~\cite{EchouffoTcheugui2012}. More recent models, including the KFRE validated by Tangri and colleagues~\cite{Tangri2016} and its UK validation by Major and colleagues~\cite{Major2019}, achieve strong external discrimination (C-statistic 0.80--0.90) in prospective cohort studies with purpose-built feature ascertainment.

The AUROC values seen on MIMIC in this study (0.48--0.58) are not comparable to those results, because the KFRE and similar models were validated on cohorts where inputs were actually measured. The comparison instead reinforces a narrower point: a model applied to harmonized data with seven imputed features is not the same as a model applied to complete data. Deployment performance depends on what the deployment environment actually records, not on what the training environment provided.

\subsection{Limitations}

The most important limitation to state plainly: the MIMIC cohort used here is the publicly available 100-patient demonstration subset released by PhysioNet for pipeline development, not a formally designed external validation cohort. It was not selected to represent a specific clinical population, it contains only 23 CKD cases, and seven of the features were entirely missing and had to be imputed from UCI training statistics. The stress-test framing in this paper is intentional and accurate. The MIMIC results show what distribution shift looks like numerically; they do not estimate performance in any real clinical deployment.

At 97 patients after filtering, calibration metrics carry wide uncertainty regardless of the above. Bootstrap confidence intervals for MIMIC ECE span as much as 0.18 for some models. Sample size requirements for precise external calibration assessment substantially exceed the 97 patients available in this stress-test cohort~\cite{Riley2024EvalPart3}. The point estimates should be read as directional, not precise.

Feature harmonization made the missingness problem worse. The seven imputed features were filled using UCI training-set statistics, so those columns carry no information specific to individual MIMIC patients. That is an inherent constraint of cross-dataset harmonization when features are domain-specific, and it cannot be corrected analytically after the fact.

Subgroup analysis was additionally limited by sample size. Diabetes and hypertension subgroups each fell below the 10-patient exclusion threshold, and the age subgroups (below 65: $n$\,=\,55; 65 and above: $n$\,=\,42) were small enough that bin-level ECE estimates carry meaningful uncertainty. Any subgroup findings should be treated as exploratory.

\subsection{Future Directions}

Three areas appear worth pursuing. First, the full MIMIC-IV database would provide several thousand patients with CKD-relevant laboratory data, allowing larger-scale calibration assessment and properly powered subgroup analysis. The extraction pipeline from this study is directly applicable to the full database.

Second, the deployment readiness framework here is intentionally simple. Criteria are binary and thresholds are set a priori without formal sample size calculations. Extending the framework to account for uncertainty in threshold exceedance, for example through bootstrap p-values for each criterion, would make the checklist scores more principled.

Third, patient-facing uncertainty display is an open design problem. Conformal prediction sets at the individual level (CKD / not-CKD / ambiguous) produce clinically interpretable labels. How clinicians and patients respond to ambiguous predictions in a real consultation, and whether explicit uncertainty communication changes treatment decisions compared to point probability outputs, has not been tested in CKD care.

%% file: sections/conclusion.tex
Five classifiers achieved perfect discrimination on the UCI CKD benchmark, then failed every stress-test criterion when applied to the MIMIC-IV demo cohort under deliberate distributional shift. Isotonic recalibration reduced internal ECE to near zero. On the demo cohort, calibration drift exceeded 0.67 for every model, conformal coverage fell from 0.967 or higher internally to 0.21--0.25, and no model scored above 4 out of 16 on the deployment readiness checklist.

The MIMIC cohort used here is an open-access 100-patient demonstration set, not a formally designed external validation dataset. The stress-test framing is intentional. Its purpose is to illustrate that internal calibration quality, however strong, does not guarantee the same reliability under prevalence shift and feature missingness. That point holds regardless of the demo cohort's limitations.

The eight-criterion framework introduced here offers one structured path for evaluating these requirements before clinical use. Each criterion has a defined threshold and a three-level scoring scheme applicable to any binary clinical prediction model. Formal external validation on a purpose-built cohort with properly measured features and an appropriate sample size is the necessary next step. The framework provides the evaluation structure. The full MIMIC-IV database provides the cohort.